\documentclass[runningheads]{llncs}
\usepackage{graphicx}
\usepackage{times}
\usepackage{tikz}
\usepackage{comment}
\usepackage{amsmath,amssymb} % define this before the line numbering.
\usepackage{xurl} % allow line breaks anywhere in URL string
\usepackage{epsfig}
\usepackage{graphicx}
\usepackage{amsmath}
\usepackage{amssymb}
\usepackage{color}
\usepackage{bm}
\usepackage{multirow}
\usepackage{diagbox}
\usepackage{array}
\usepackage{booktabs}
\usepackage{nth}
\usepackage{nicefrac}
\usepackage{relsize}
\usepackage{xspace} 
\usepackage{bm}
\usepackage{enumerate}
\usepackage{adjustbox}
\usepackage{enumitem}
\usepackage{array}
\usepackage{cuted}
\usepackage{capt-of}
\usepackage{authblk}
\usepackage[symbol]{footmisc}
\renewcommand{\thefootnote}{\fnsymbol{footnote}}

\newcolumntype{P}[1]{>{\centering\arraybackslash}p{#1}}
\usepackage[pagebackref=true,breaklinks=true,colorlinks,bookmarks=false]{hyperref}

\usepackage[accsupp]{axessibility}
\newcommand\blfootnote[1]{%
  \begingroup
  \renewcommand\thefootnote{}\footnote{#1}%
  \addtocounter{footnote}{-1}%
  \endgroup
}

\newcommand{\ie}{i.e.~}
\newcommand{\etal}{et al.~}
\newcommand{\eg}{e.g.~}
\newcommand{\mytilde}{\raise.17ex\hbox{$\scriptstyle\mathtt{\sim}$}}

\begin{document}
\pagestyle{headings}
\mainmatter

\title{Free-Viewpoint RGB-D Human Performance \\Capture and Rendering} 

\titlerunning{Free-Viewpoint RGB-D Human Performance Capture and Rendering} 
\author{Phong Nguyen-Ha \inst{1*} \and
Nikolaos Sarafianos \inst{2}\and \\
Christoph Lassner \inst{2} \and Janne Heikkil\"a \inst{1} \and
Tony Tung \inst{2}}
\authorrunning{P. Nguyen et al.}
\institute{Center for Machine Vision and Signal Analysis, University of Oulu, Finland
\and
Meta Reality Labs Research, Sausalito\\ \href{https://www.phongnhhn.info/HVS_Net}{https://www.phongnhhn.info/HVS\_Net}} 

\maketitle

\begin{abstract}
\noindent
Capturing and faithfully rendering photorealistic humans from novel views is a fundamental problem for AR/VR applications. While prior work has shown impressive performance capture results in laboratory settings, it is non-trivial to achieve casual free-viewpoint human capture and rendering for unseen identities with high fidelity, especially for facial expressions, hands, and clothes. To tackle these challenges we introduce a novel view synthesis framework that generates realistic renders from unseen views of any human captured from a single-view and sparse RGB-D sensor, similar to a low-cost depth camera, and without actor-specific models.
We propose an architecture to create dense feature maps in novel views obtained by sphere-based neural rendering, and create complete renders using a global context inpainting model.
Additionally, an enhancer network leverages the overall fidelity, even in occluded areas from the original view, producing crisp renders with fine details.
We show that our method generates high-quality novel views of synthetic and real human actors given a single-stream, sparse RGB-D input.
It generalizes to unseen identities, and new poses and faithfully reconstructs facial expressions.
Our approach outperforms prior view synthesis methods and is robust to different levels of depth sparsity.
\end{abstract}

\section{Introduction}
\label{intro}
\blfootnote{*This work was conducted during an internship at Meta Reality Labs Research.}
Novel view synthesis of rigid objects or dynamic scenes has been a very active topic of research recently with impressive results across various tasks~\cite{mildenhall2020nerf,RGBDNet,NonRigidNERF}.
However, synthesizing novel views of humans in motion requires methods to handle dynamic scenes with various deformations which is a challenging task~\cite{NonRigidNERF,yoon2020dynamic}; especially in those regions with fine details such as the face or the clothes~\cite{2021narf,ani_nerf,wang2022animatable,H_nerf}.
In addition, prior work usually relies on a large amount of cameras~\cite{BansalCVPR20,mildenhall2020nerf}, expensive capture setups~\cite{neural_body}, or inference time on the order of several minutes per frame.
This work aims to tackle these challenges using a compact, yet effective formulation. 

We propose a novel \textbf{H}uman \textbf{V}iew \textbf{S}ynthesis \textbf{Net}work (\textbf{HVS-Net}) that generates high-fidelity rendered images of clothed humans using a commodity RGB-D sensor.
The challenging requirements that we impose are: i)~generalization to new subjects at test-time as opposed to models trained per subject, ii)~the ability to handle dynamic behavior of humans in unseen poses as opposed to animating humans using the same poses seen at training, iii)~the ability to handle occlusions (either from objects or self-occlusion), iv)~capturing facial expressions and v)~the generation of high-fidelity images in a live setup given a single-stream, sparse RGB-D input (similar to a low-cost, off-the-shelf depth camera). 

\setlength{\belowcaptionskip}{-13pt}
\begin{figure}[t]
\centering
  \includegraphics[width=0.99\linewidth]{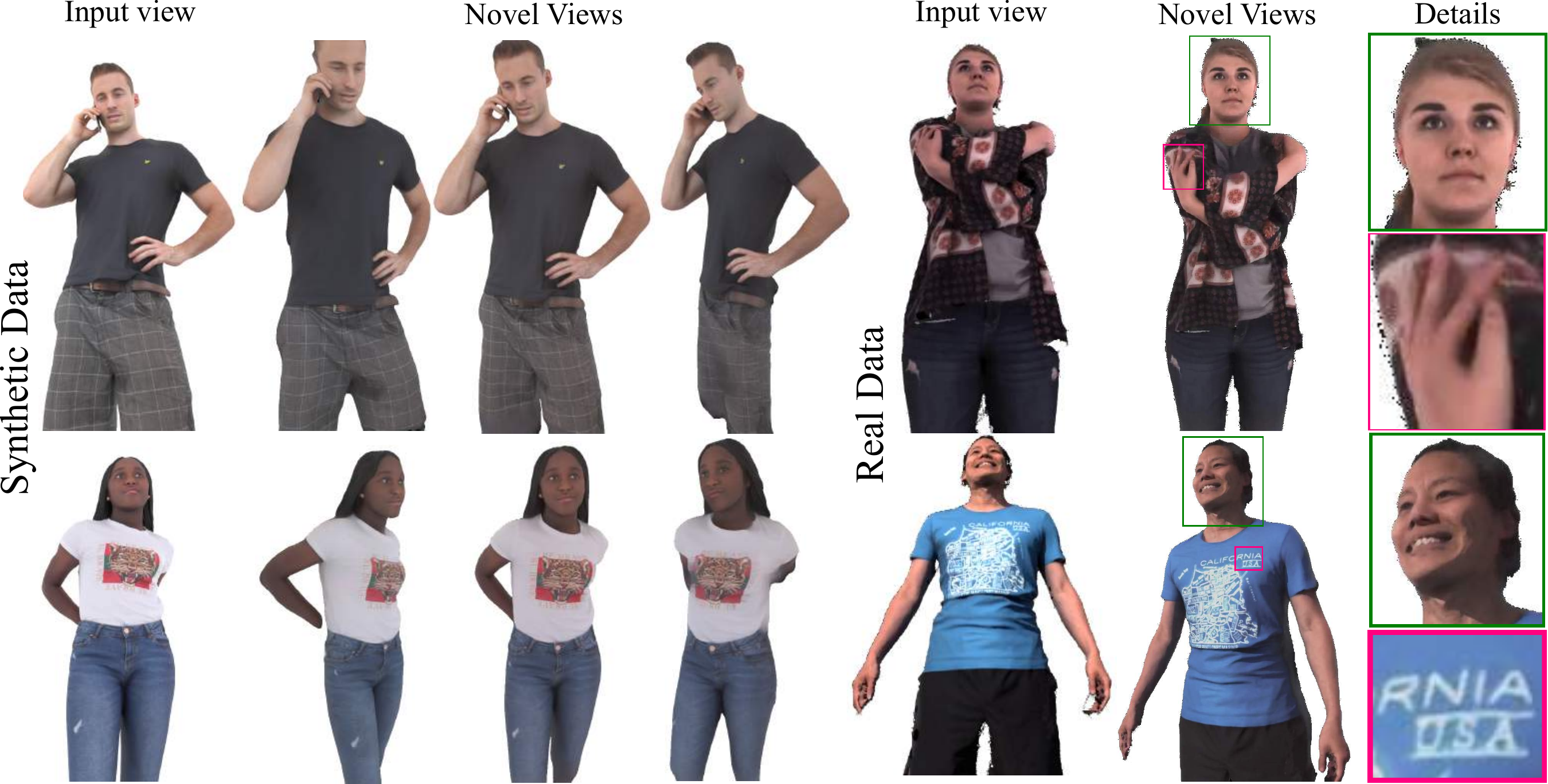}
\captionof{figure}{\textit{Overview.} We present a Human View Synthesis model that predicts novel views of humans from a single-view, sparse RGB-D input. Our method renders high quality novel views of both, synthetic and real humans at 1K resolution without per-subject fine tuning.}
\label{fig_teaser}
\end{figure}

HVS-Net takes as input a single, sparse RGB-D image of the upper body of a human and a target camera pose and generates a high-resolution rendering from the target viewpoint (see Fig.~\ref{fig_teaser}).
The first key differentiating factor of our proposed approach compared to previous approaches is that we utilize depth as an additional input stream.
While the input depth is sparse and noisy it still enables us to utilize the information seen in the input view and hence simplifying the synthesis of novel views.
To account for the sparseness of the input, we opted for a sphere-based neural renderer that uses a learnable radius to create a denser, warped image compared to simply performing geometry warping from one view to the other. 
When combined with an encoder-decoder architecture and trained end-to-end, our approach is able to synthesize novel views of unseen individuals and to in-paint areas that are not visible from the main input view.
However, we observed that while this approach works well with minimal occlusions it has a hard time generating high-quality renderings when there are severe occlusions, either from the person moving their hands in front of their body or if they're holding various objects.
Thus, we propose to utilize a single additional occlusion-free image and warp it to the target novel view by establishing accurate dense correspondences between the two inputs.
A compact network can be used for this purpose, which is sufficient to refine the final result and generate the output prediction. 
We train the entire pipeline end-to-end using photometric losses between the generated and ground-truth pair of images. In addition, we use stereoscopic rendering to encourage view-consistent results between close-by viewpoints.
To train HVS-Net, we rely on high-quality synthetic scans of humans that we animated and rendered from various views.
A key finding of our work is that it generalizes very well to real data captured by a 3dMD scanner system with a level of detail in the face or the clothes that are not seen in prior works~\cite{kwon2021neural,kwon2020rotationally,neural_body}.
In summary, the contributions of this work are: 
\begin{itemize}[leftmargin=*]
    \item A robust sphere-based synthesis network that generalizes to multiple identities without per-human optimization.
    \item A refinement module that enhances the self-occluded regions of the initial estimated novel views. This is accomplished by introducing a novel yet simple approach to establish dense surface correspondences for the clothed human body that addresses key limitations of DensePose which is usually used for this task.
    \item State-of-the-art results on dynamic humans wearing various clothes, or accessories and with a variety of facial expressions of both, synthetic and real-captured data.
\end{itemize}

\section{Related Work}\label{related_work}

View synthesis for dynamic scenes, in particular for humans, is a well-established field that provides the basis for this work. Our approach builds on ideas from point-based rendering, warping, and image-based representations.

\noindent \textbf{View Synthesis.}
For a survey of early image-based rendering methods, we refer to \cite{shum2000,SOTA_report}.
One of the first methods to work with video in this field is presented in \cite{carranza2003} and uses a pre-recorded performance in a multi-view capturing setup to create the free-viewpoint illusion.
Zitnick \etal~\cite{Zitnick2004}~similarly use a multi-view capture setup for viewpoint interpolation. 
These approaches interpolate between recorded images or videos. Ballan \etal~\cite{Ballan2010} coin the term `video-based rendering': they use it to interpolate between hand-held camera views of performances.
The strong generative capabilities of neural networks enable further extrapolation and relaxation of constraints~\cite{Flynn2016,Huang2018DeepVV,Kalantari2016,Meshry2019}.
Zhou \etal~\cite{Zhou2018} introduce Multi-Plane Images (MPIs) for viewpoint synthesis and use a model to predict them from low-baseline stereo input and \cite{Flynn2019,Srinivasan2019} improve over the original baseline and additionally work with camera arrays and light fields.
Broxton \etal~\cite{Broxton2020}~extend the idea to layered, dynamic meshes for immersive video experiences whereas Bansal \etal~\cite{BansalCVPR20}~use free camera viewpoints, but multiple cameras.
With even stronger deep neural network priors, \cite{Wiles_2020_CVPR}~performs viewpoint extrapolation from a single view, but for static scenes, whereas~\cite{NonRigidNERF,yoon2020dynamic} can work with a single view in dynamic settings with limited motion.
Bemana \etal \cite{xfields} works in static settings but predicts not only the radiance field but also lighting given varying illumination data.
Chibane \etal~\cite{Chibane2021} trade instant depth predictions and synthesis for the requirement of multiple images.
Alternatively, volumetric representations~\cite{Lombardi:2019,Lombardi2021} can also being utilized for capturing dynamic scenes.
All these works require significant computation time for optimization, multiple views or offline processing for the entire sequence.

\noindent \textbf{3D \& 4D Performance Capture.}
While the aforementioned works are usually scene-agnostic, employing prior knowledge can help in the viewpoint extrapolation task: this has been well explored in the area of 3D \& 4D Human Performance Capture. A great overview of the development of the \emph{Virtualized Reality} system developed at CMU in the 90s is presented in~\cite{virtualized-reality}.
It is one of the first such systems and uses multiple cameras for full 4D capture.
Starting from this work, there is a continuous line of work refining and improving over multi-view capture of human performances~\cite{perfcap,collet2015,li2012,Zitnick2004}.
Relightables~\cite{relightables} uses again a multi-camera system and adds controlled lighting to the capture set up, so that the resulting reconstructed performances can be replayed in new lighting conditions.
The authors of~\cite{panoptic}~take a different route: they find a way to use bundle adjustment for triangulation of tracked 3D points and obtain results with sub-frame time accuracy.
Broxton \etal~\cite{Broxton2020} is one of the latest systems for general-purpose view interpolation and uses a multi-view capture system to create a layered mesh representation of the scene.
Many recent works apply neural radiance fields~\cite{mildenhall2020nerf,xie2021neuralfield} to render humans at novel views. Li \etal~\cite{Li2021}~use a similar multi-view capture system to train a dynamic Neural Radiance Field.
Kwon \etal~\cite{kwon2021neural} learn generalizable neural radiance fields based on a parametric human body model to perform novel view synthesis. However, this method fails to render high-quality cloth details or facial expressions of the human.
Both of these systems use multiple cameras and are unable to transmit performance in real-time.
Given multi-view input frames or videos, recent works on rendering animate humans from novel views show impressive results~\cite{2021narf,ani_nerf,neural_body,H_nerf}.
However such methods can be prohibitively expensive to run (\cite{2021narf} runs at 1 minute/frame) and cannot generalize to unseen humans but instead create a dedicated model for each human that they need to render.

\noindent \textbf{Human View Synthesis using RGB-D.}
A few methods have been published recently that tackle similar scenarios: LookingGood~\cite{LookingGood} re-renders novel viewpoints of a captured individual given a single RGB-D input.
However, their capture setup produces dense geometry which makes this a comparatively easy task: the target views do not deviate significantly from the input views.
A recent approach~\cite{VC} uses a frontal input view and a large number of calibration images to extrapolate novel views. This method relies on a keypoint estimator to warp the selected calibrated image to the target pose, which leads to unrealistic results for hands, occluded limbs, or for large body shapes.

\noindent \textbf{Point-based Rendering.} We assume a single input RGB-D sensor as a data source for our method.
This naturally allows us to work with the depth data in a point-cloud format.
To use this for end-to-end optimization, we build on top of ideas from differentiable point cloud rendering.
Some of the first methods rendered point clouds by blending discrete samples using local blurring kernels: \cite{insafutdinov2018unsupervised,lin2018learning,roveri2018network}.
Using the differentiable point cloud rendering together with convolutional neural networks naturally enables the use of latent features and a deferred rendering layer, which has been explored in~\cite{lassner2021pulsar,Wiles_2020_CVPR}.
Recent works on point-based rendering~\cite{aliev2020neural,KPLD21} use a point renderer implemented in OpenGL, then use a neural network image space to create novel views.
Ruckert \etal~\cite{ruckert2021adop} use purely pixel-sized points and finite differences for optimization.
We are directly building on these methods and use the Pulsar renderer~\cite{lassner2021pulsar} in our method together with an additional model to improve the point cloud density.

\noindent \textbf{Warping Representations.} To correctly render occluded regions, we warp the respective image regions from an unoccluded posture to the required posture.
Debevec \etal~\cite{Debevec1998}~is one of the first methods to use ``projective texture-mapping'' for view synthesis.
Chaurasia \etal~\cite{chaurasia2013}~uses depth synthesis and local warps to improve over image-based rendering.
The authors of~\cite{deepwarp}~take view synthesis through warping to its extreme: they solely use warps to create novel views or synthesize gaze.
Recent methods~\cite{RGBDNet,Riegler2020FVS,Thies2020} use 3D proxies together with warping and a CNN to generate novel views.
All these methods require either creation of an explicit 3D proxy first, or use of image-based rendering.
Instead, we use the dynamic per-frame point cloud together with a pre-captured, unoccluded image to warp necessary information into the target view during online processing.

\begin{figure}[t]
\centering
  \includegraphics[width=0.8\linewidth]{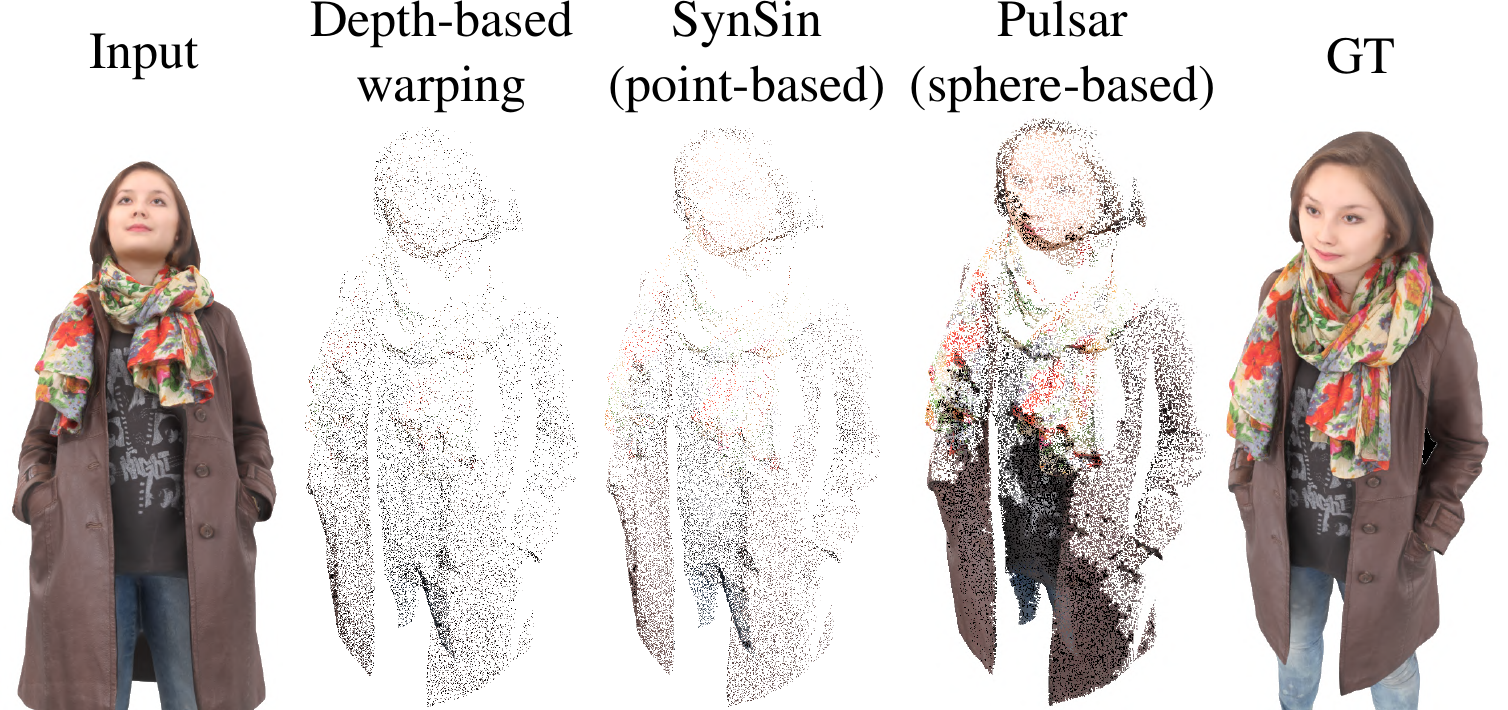}
  \caption{\textit{Comparison of 3D point cloud transformations.} From a single RGB-D input, we obtain the warped image using: a depth-based warping transformation~\cite{Forward_warping,LookingGood}, the neural point-based renderer SynSin~\cite{Wiles_2020_CVPR} and the neural sphere-based Pulsar renderer~\cite{lassner2021pulsar}. The novel image warped by Pulsar is significantly denser.}
\label{fig_warp}
\end{figure}

\section{HVS-Net Methodology}\label{method}

The goal of our method is to create realistic novel views of a human captured by a single RGB-D sensor (with sparse depth, similar to a low-cost RGB-D camera), as faithful and fast as possible.
We assume that the camera parameterization of the view to generate is known.
Still, this poses several challenges: 1) the information we are working with is incomplete, since not all regions that are visible from the novel view can be observed by the RGB-D sensor; 2) occlusion adds additional regions with unknown information; 3) even the pixels that are correctly observed by the original sensor are sparse and exhibit holes when viewed from a different angle.
We tackle the aforementioned problems using an end-to-end trainable neural network with two components.

First, given an RGB-D image parameterized as its two components RGB $I_v$ and sparse depth $D_v$ taken from the input view $v$, a sphere-based view synthesis model \textit{S} produces dense features of the target view and renders the resulting RGB image from the target camera view using a global context inpainting network $G$ (see Sec.~\ref{SVS}).
However, this first network can not fully resolve all occlusions: information from fully occluded regions is missing (e.g., rendering a pattern on a T-shirt that is occluded by a hand).
To account for such cases, we optionally extend our model with an enhancer module $E$ (see Sec.~\ref{enhance_net}).
It uses information from an unoccluded snapshot of the same person, estimates the dense correspondences between the predicted novel view and occlusion-free input view, and then refine the predicted result.

\subsection{Sphere-based View Synthesis} \label{SVS}

The goal of this first part of our pipeline is to render a sparse RGB-D view of a human as faithfully as possible from a different perspective.
Of the aforementioned artifacts, it can mostly deal with the inherent sparsity of spheres caused due to the depth foreshortening: from a single viewpoint in two neighboring pixels, we only get a signal at their two respective depths---no matter how much they differ.
This means that for every two pixels that have a large difference in depth and are seen from the side, large gaps occur.
For rendering human subjects, these ``gaps'' are of limited size, and we can address the problem to a certain extent by using a sphere-based renderer for view synthesis.

\begin{figure*}[t]
\centering
  \includegraphics[width=1.0\linewidth]{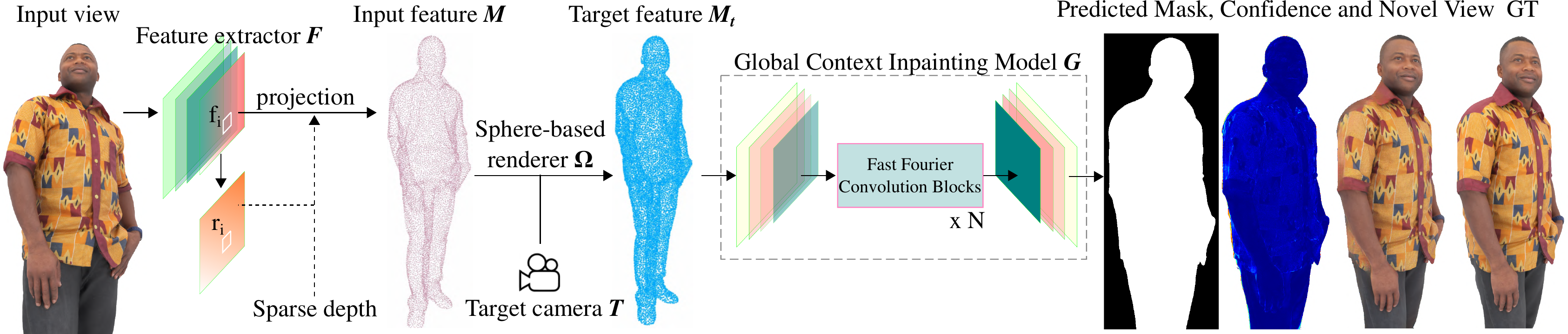}
  \caption{\textit{Sphere-based view synthesis network architecture.} The feature predictor $F$ learns radius and feature vectors of the sphere set $S$. We then use the sphere-based differentiable renderer $\Omega$ to densify the learned input features $M$ and warp them to the target camera $T$. The projected features $M_t$ are passed through the global context inpainting module $G$ to generate the foreground mask, confidence map and novel image. Brighter confidence map colors indicate lower confidence.}
\label{fig_SVS}
\end{figure*}

\noindent\textbf{Sphere-based renderer.}
Given the depth of every pixel from the original viewpoint as well as the camera parameters, these points can naturally be projected into a novel view.
This makes the use of depth-based warping or of a differentiable point- or sphere-renderer a natural choice for the first step in the development of the view synthesis model.
The better this renderer can transform the initial information into the novel view, the better; this projection step is automatically correct (except for sensor noise) and not subject to training errors.

In Fig.~\ref{fig_warp}, we compare the density of the warped images from a single sparse RGB-D input using three different methods: depth-based warping~\cite{Forward_warping}, point-based rendering~\cite{Wiles_2020_CVPR} and sphere-based rendering~\cite{lassner2021pulsar}.
Depth based warping~\cite{Forward_warping} represents the RGD-D input as a set of pixel-sized 3D points and thus, the correctly projected pixels in the novel view are very sensitive to the density of the input view.
The widely-used differentiable point-based renderer~\cite{Wiles_2020_CVPR} introduces a global radius-per-point parameter which allows to produce a somewhat denser images.
Since it uses the same radius for all points, this comes, however, with a trade-off: if the radius is selected too large, details in dense regions of the input image are lost; if the radius is selected too small, the resulting images get sparser in sparse regions.
The recently introduced, sphere-based Pulsar renderer~\cite{lassner2021pulsar} not only provides the option to use a per-sphere radius parameter, but it also provides gradients for these radiuses, which enables us to set them dynamically.
As depicted in Fig.~\ref{fig_warp}, this allows us to produce denser images compared to the other methods.
Fig.~\ref{fig_SVS} shows an overview of the overall architecture of our method.
In a first step, we use a shallow set of convolutional layers $F$ to encode the input image $I_v$ to a $d$-dimensional feature map $M = F(I_v)$.
From this feature map, we create a sphere representation that can be rendered using the Pulsar renderer.
This means that we have to find position $p_i$, feature vector $f_i$, and radius $r_i$ for every sphere $i\in{1,..,N}$ when using $N$ spheres (for further details about the rendering step, we refer to ~\cite{lassner2021pulsar}).
The sphere positions $p_i$ can trivially be inferred from camera parameters, pixel index and depth for each of the pixels.
We choose the features $f_i$ as the values of $M$ at the respective pixel position; we infer $r_i$ by passing $M$ to another convolution layer with a sigmoid activation function to bound its range. This leads to an as-dense-as-possible projection of features into the target view, which is the basis for the following steps. 

\noindent\textbf{Global context inpainiting model.}
Next, the projected features are converted to the final image.
This remains a challenging problem since several ``gaps'' in the re-projected feature images $M_t$ cannot be avoided.
To address this, we design an efficient encoder-decoder-based inpainting model $G$ to produce the final renders.
The encoding bottleneck severely increases the receptive field size of the model, which in turn allows it to correctly fill in more of the missing information.
Additionally, we employ a series of Fast Fourier Convolutions (FFC)~\cite{FFC} to take into account the image-wide receptive field. The model is able to hallucinate missing pixels much more accurately compared to regular convolution layers~\cite{LAMA}. 

\noindent\textbf{Photometric Losses}.
The sphere-based view synthesis network $S$ not only predicts an RGB image $I_p$ of the target view, but also a foreground mask $I_m$ and a confidence map $I_c$ which can be used for compositing and error correction, respectively.
We then multiply the predicted foreground mask and confidence map with the predicted novel image: $I_p = I_p * I_m * I_c$. However, an imperfect mask $I_m$
may bias the network towards unimportant areas. Therefore, we predict a confidence mask $I_c$ as a side-product of the G network to dynamically assign less weight to “easy” pixels, whereas “hard” pixels get higher importance~\cite{LookingGood}.

All of the aforementioned model components are trained end-to-end using the photometric loss $\mathcal{L}_{photo}$, which is defined as: $\mathcal{L}_{photo} = \mathcal{L}_{i} + \mathcal{L}_{m}$.
$\mathcal{L}_{i}$ is the combination of an $\ell_1$, perceptual~\cite{vggLoss} and hinge GAN~\cite{GAN} loss between the estimated new view $I_{p}$ and the ground-truth image $I_{GT}$.
$\mathcal{L}_{m}$ is the binary cross-entropy loss between the predicted and ground-truth foreground mask.
We found that this loss encourages the model to predict sharp contours in the novel image.
The two losses lead to high-quality reconstruction results for single images.
However, we note that stereoscopic rendering of novel views requires matching left and right images for both views. Whereas the above losses lead to \emph{plausible} reconstructions, they do not necessarily lead to sufficiently consistent reconstructions for close-by viewpoints.
We found a two-step strategy to address this issue: 
1) Instead of predicting a novel image of a single viewpoint, we train the model to predict two nearby novel views.
To obtain perfectly consistent depth between both views, we use the warping operator $W$ from ~\cite{STN} to warp the predicted image and the depth from one to the nearby paired viewpoint.
2) In the second step, we define a multi-view consistency loss $\mathcal{L}_{c}$ as:
\begin{equation}
\label{eqn_consistent_loss}
\mathcal{L}_{c} = ||I_{p}^L - W(I_{p}^R)||_1,
\end{equation}
where $I_{p}^L$ and $I_{p}^R$ are predicted left and right novel views.
With this, we define the photometric loss as follows:
\begin{equation}
\label{eqn_photo}
\mathcal{L}_{photo} = \mathcal{L}_{i} + 0.5 \times \mathcal{L}_{m} + 0.5 \times \mathcal{L}_{c}.
\end{equation}

\subsection{Handling Occlusions} \label{enhance_net}

\begin{figure}[t]
\centering
  \includegraphics[width=0.8\linewidth]{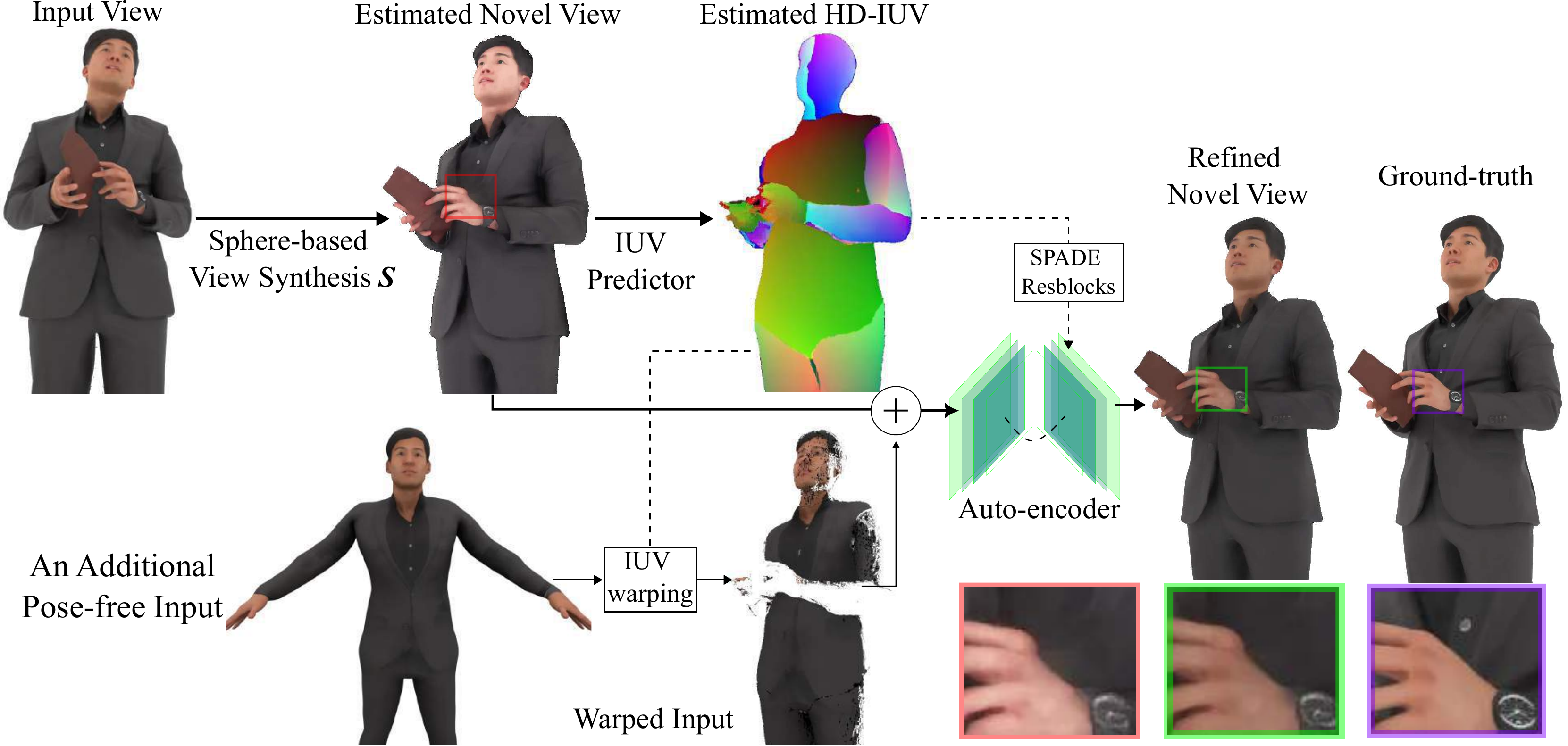}
  \caption{\textit{IUV-based image refinement.} Using an additional occlusion-free input, we refine the initial estimated novel view by training the Enhancer network $E$. We infer the dense correspondences of both, predicted novel view and occlusion-free image, using a novel \textit{HD-IUV} module. The occlusion-free image is warped to the target view and then refined by an auto-encoder. The refined novel view shows crisper results on the occluded area compared to the initially estimated render.}
\label{fig_ENet}
\end{figure}

The sphere-based view synthesis network $S$ predicts plausible novel views with high quality.
However, if the person is holding an object such as a wallet (c.t. Fig.~\ref{fig_ENet}) or if their hands are obstructing large parts of their torso, then the warped transformation will result in missing points in this region (as discussed in Fig~\ref{fig_warp}).
This leads to low-fidelity texture estimates for those occluded regions when performing novel view synthesis with a target camera that is not close to the input view.
Hence, to further enhance the quality of the novel views, we introduce two additional modules: \textit{i)} an \textit{HD-IUV} predictor $D$ to predict dense correspondences between an RGB image (render of a human) and the 3D surface of a human body template, and \textit{ii)} a refinement module $R$ to warp an additional occlusion-free input (\eg, a selfie in a practical application) to the target camera and enhance the initial estimated novel view to tackling the self-occlusion issue. 

\begin{figure}[t]
\centering
  \includegraphics[width=0.9\linewidth]{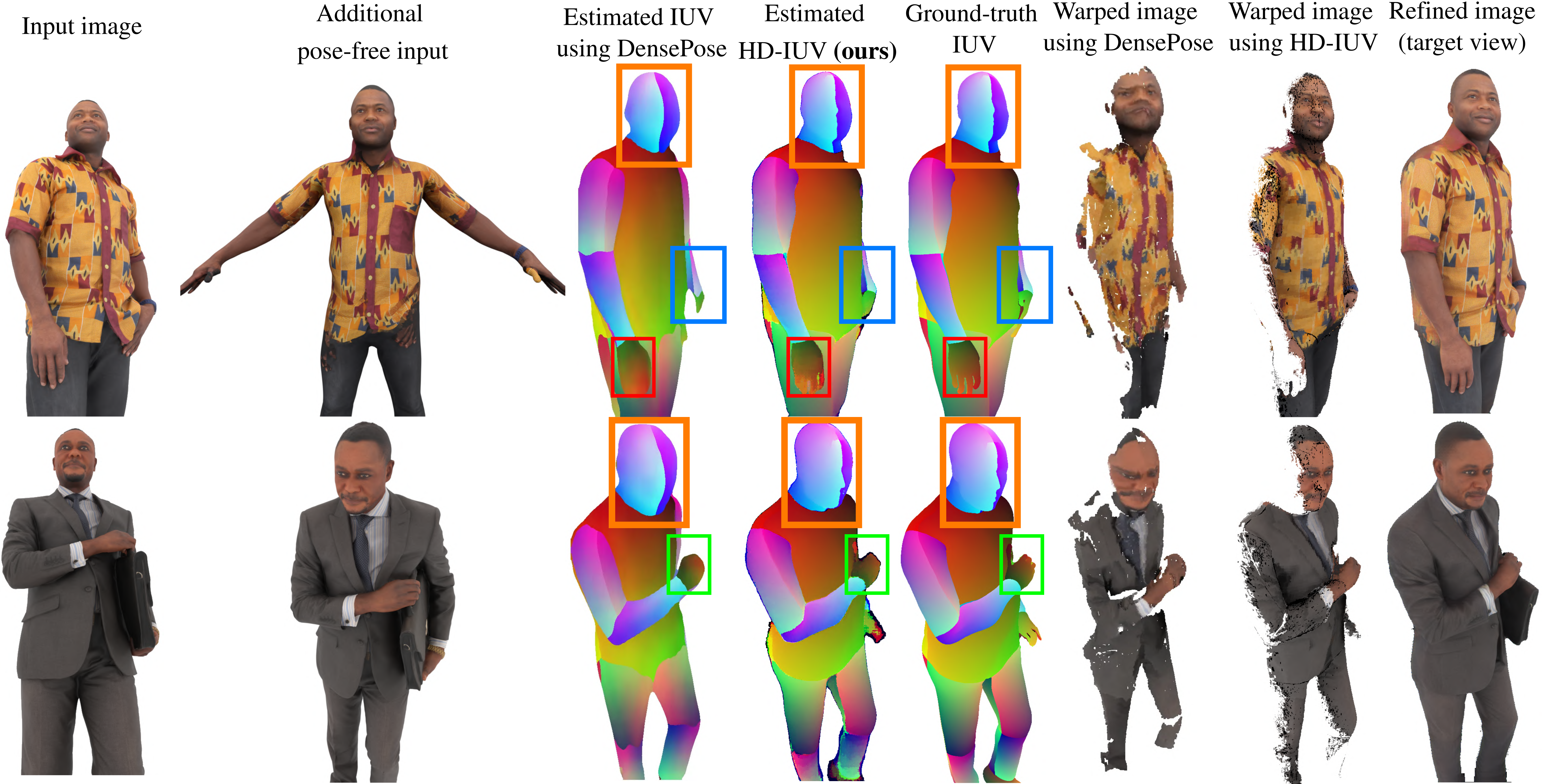}
  \caption{\textit{Dense correspondence visualization.} Texture warping with DensePose results in inaccurate and distorted images in the target view due to incorrect IUV estimates (enhanced by the fact that it targets the naked body). Our proposed HD-IUV representation covers the human body including clothing, captures facial and hand details with high accuracy, and results in less distorted renderings in the target view. We stack this warped image with the initially estimated target-view synthesized image and provide it as input for the Enhancer network to obtain the final results.}
\label{fig_qual_iuv}
\end{figure}

\noindent \textbf{HD-IUV Predictor $D$}. We first estimate a representation that maps an RGB image of a human to the 3D surface of a body template~\cite{ianina2022bodymap,neverova2018dense,neverova2020continuous,tan2021humangps}.
One could use DensePose~\cite{neverova2018dense} for this task but the estimated IUV (where I reflects the body part) predictions cover only the naked body instead of the clothed human and are inaccurate as they are trained based on sparse and noisy human annotations.
Instead, we build our own IUV predictor and train it on synthetic data for which we can obtain accurate ground-truth correspondences.
With pairs of synthetic RGB images and ground-truth dense surface correspondences, we train a UNet-like network that provides dense surface (\ie IUV) estimates \emph{for each pixel} of the clothed human body. 
For each pixel $p$ in the foreground image, we predict 3-channeled (RGB) color $p'$ which represents the correspondence (the colors in such a representation are unique which makes subsequent warping easy).
Thus, we treat the whole problem as a multi-task classification problem where each task (predictions for the I, U, and V channels) is trained with the following set of losses: a) multi-task classification loss for each of the 3 channels (per-pixel classification label) and b) silhouette loss.
In Fig.~\ref{fig_qual_iuv} we show that, unlike DensePose, the proposed HD-IUV module accurately establishes fine level correspondences for the face and hand regions while capturing the whole clothed human and thus making it applicable for such applications. 
Once this model is pre-trained, we merge it with the rest of the pipeline and continue the training procedure by using the initially estimated novel view $I_p$ ad an input to an encoder-decoder architecture that contains three prediction heads (for the I, U, and V channels). An in-depth discussion on the data generation, network design, and training is provided in the supplementary material. 

\noindent \textbf{Warping Representations and View Refinement}.
The predicted \textit{HD-IUV} in isolation would not be useful for the task of human view synthesis.
However, when used along with the occlusion-free RGB input, it allows us to warp all visible pixels to the human in the target camera $T$ and obtain a partial warped image $I_w$.
For real applications this occlusion-free input can be a selfie image---there are no specific requirements to the body pose for the image.
In Fig.~\ref{fig_qual_iuv} we compare DensePose results with the proposed HD-IUV module.
DensePose clearly produces less accurate and more distorted textures.

In the next step, we stack $I_p$ and $I_w$ and pass the resulting tensor to a refinement module.
This module addresses two key details: a) it learns to be robust to artifacts that are originating either from the occluded regions of the initially synthesized novel view as well as texture artifacts that might appear due to the fact that we rely on HD-IUV dense correspondences for warping and b) it is capable of synthesizing crisper results in the occluded regions as it relies on both the initially synthesized image as well as the warped image to the target view based on HD-IUV.
The refinement module is trained using the photometric loss $\mathcal{L}_{photo}$ between the refined novel images and ground truths.
All details regarding  training and image warping, as well as the full network architecture, can be found in the supplementary material.

\section{Experiments} \label{Exp}
\noindent \textbf{Datasets}.
The proposed approach is trained solely on synthetic data and evaluated quantitatively and qualitatively on both, synthetic and real data.
For training, we use the RenderPeople dataset~\cite{renderppl}, which has been used extensively~\cite{alldieck2019tex2shape,bhatnagar2019mgn,3Dhumantexturesynthesis,huang2020arch,palafox2021spams,agora,tan2021humangps} for human reconstruction and generation tasks. 
Overall, we use a subset of 1000 watertight meshes of people wearing a variety of garments and in some cases holding objects such as mugs, bags or mobile phones.
Whereas this covers a variety of personal appearances and object interaction, all of these meshes are static---the coverage of the pose space is lacking.
Hence, we augment the dataset by introducing additional pose variations: we perform non-rigid registration for all meshes, rig them for animation and use a set of pre-defined motions to animate them.
With this set of meshes \emph{and} animations, we are able to assemble a set of high-quality ground-truth RGB-D renders as well as their corresponding IUV maps for 25 views per frame using Blender.
We use a 90/10 train/test split based on identities to evaluate whether our model can generalize well to unseen individuals.

In addition to the synthetic test set, we also assemble a real-world test dataset consisting of 3dMD 4D scans of people in motion.
The 3dMD 4D scanner is a full-body scanner that captures unregistered volumetric point clouds at 60Hz.
We use this dataset solely for testing to investigate how well our method handles the domain gap between synthetic and real data.
The 3dMD data does not include object interactions, but is generally noisier and has complex facial expressions.
To summarize: our training set comprises 950 static scans in their original pose and \mytilde10000 posed scans after animation.
Our test set includes 50 static unseen identities along with 1000 animated renders and 3000 frames of two humans captured with a 3dMD full-body scanner.

\noindent \textbf{Novel Viewpoint Range}. We assume a scenario with a camera viewpoint at a lower level in front of a person (\eg, the camera sitting on a desk in front of the user). This is a more challenging scenario than LookingGood~\cite{LookingGood} or Volumetric Capture~\cite{VC} use, but also a realistic one: it corresponds to everyday video conference settings. At the same time, the target camera is moving freely in the frontal hemisphere around the person (Pitch \& Roll: \([-45^o, 45^o]\), \(L_x:[-1.8m, 1.8m]\), \(L_y:[1.8m, 2.7m]\), \(L_z:[0.1m, 2.7m]\) in a Blender coordinate system). Thus, the viewpoint range is significantly larger per input view than in prior work.

\noindent \textbf{Baselines}.
In this evaluation, we compare our approach to two novel view synthesis baselines by comparing the performance in generating single, novel-view RGB images.
To evaluate the generalization of HVS-Net, we compare it with LookingGood~\cite{LookingGood}.
Since there is no available source code of LookingGood, we reimplemented the method for this comparison and validated in various synthetic and real-world settings that this implementation is qualitatively equivalent to what is reported in the original paper (we include comparison images in the supp.mat.). 
We followed the stereo set up of LookingGood and use a dense depth map to predict the novel views.
Furthermore, we compare HVS-Net with the recently proposed view synthesis method SynSin~\cite{Wiles_2020_CVPR}, which estimates monocular depth using a depth predictor.
To create fair evaluation conditions, we replace this depth predictor and either provide dense or sparse depth maps as inputs directly.
While there are several recently proposed methods in the topic of human-view synthesis; almost all are relying on either proprietary data captured in lab environments~\cite{VC}, multi-view input streams~\cite{kwon2021neural,Li2021,neural_body,H_nerf} and most importantly none of these works can generalize to new human identities (or for the case of Neural Body~\cite{neural_body} not even new poses) at testing time which our proposed HVS-Net can accomplish. Furthermore, inferring new views in a real-time manner is far from solved for most these works.
In contrast, our method focuses more on a practical approach of single view synthesis, aiming to generalize to new identities and unseen poses while being fast at inference time. Hence we stick to performing quantitative comparisons against LookingGood~\cite{LookingGood} and SynSin~\cite{Wiles_2020_CVPR} and we do not compare it with NeRF-based approaches~\cite{kwon2021neural,Li2021,neural_body,H_nerf} as such comparisons are not applicable. 

\begin{figure}[t]
  \centering
  \includegraphics[width=0.85\linewidth]{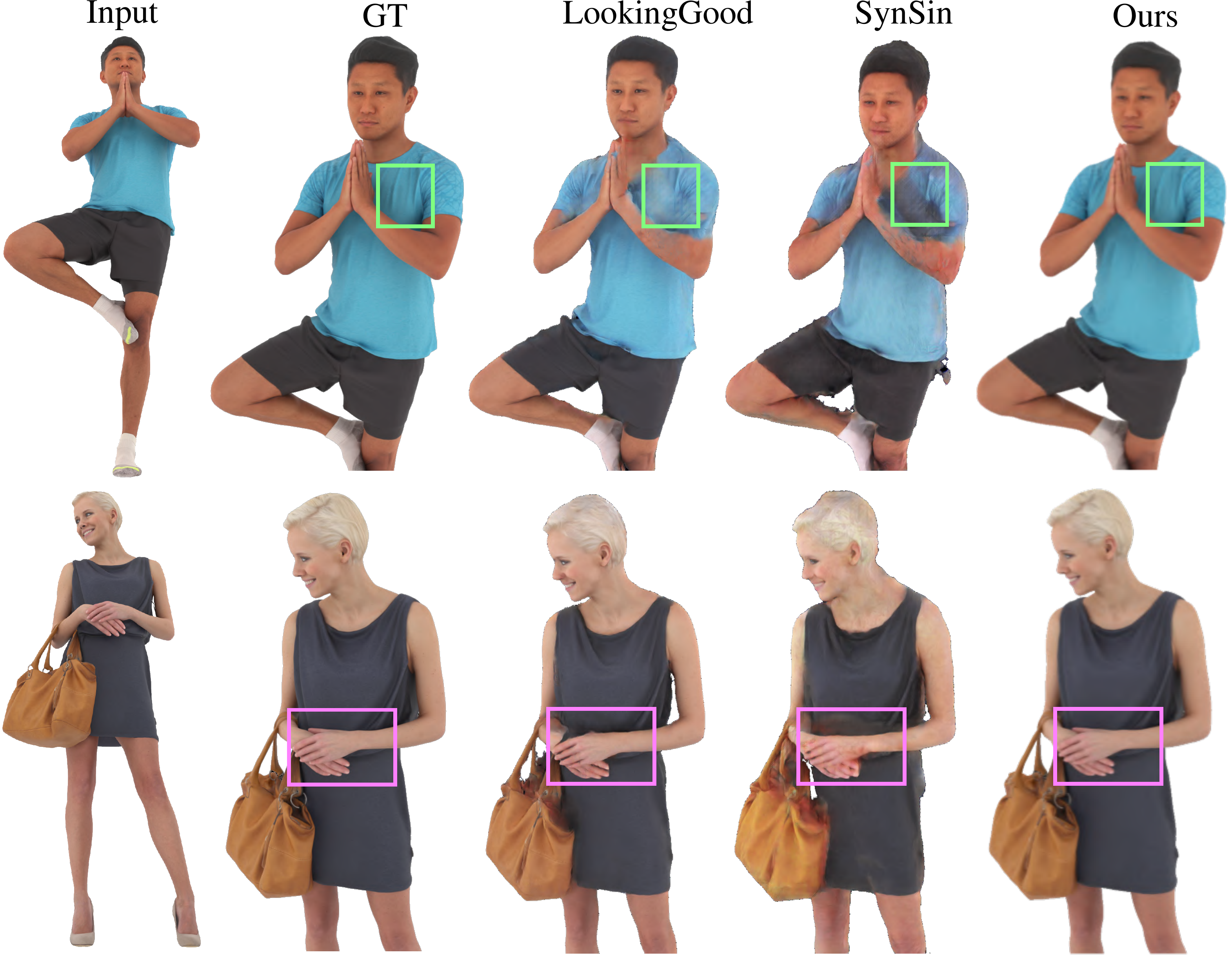}
  \caption{\textit{Qualitative comparison.} Examples of generated novel views by HVS-Net and state-of-the-art methods on the test set of the RenderPeople~\cite{renderppl} dataset. As opposed to all other methods, LookingGood~\cite{LookingGood} uses dense input depth.}
\label{fig_qual}
\end{figure}

\begin{table}[t]
\centering
\resizebox{\linewidth}{!}{
\begin{tabular}{@{}lccccccccc@{}}
\toprule
\multirow{2}{*}{Method} &
  \multicolumn{3}{c}{RenderPeople (static)} &
  \multicolumn{3}{c}{RenderPeople (animated)} &
  \multicolumn{3}{c}{Real 3dMD Data} \\ \cmidrule(l){2-4} \cmidrule(l){5-7} \cmidrule(l){8-10}
            & LPIPS$\downarrow$         & SSIM$\uparrow$  & PSNR$\uparrow$           & LPIPS$\downarrow$ & SSIM$\uparrow$           & PSNR$\uparrow$  & LPIPS$\downarrow$ & SSIM$\uparrow$  & PSNR$\uparrow$  \\ \midrule
$\text{LookingGood}^{\dagger}$~\cite{LookingGood} & 0.24          & 0.925 & 25.32          & 0.25  & 0.912          & 24.53 & 0.29  & 0.863 & 25.12 \\
$\text{SynSin}^{\dagger}$~\cite{Wiles_2020_CVPR}      & 0.31          & 0.851 & 24.18          & 0.35  & 0.937          & 23.64 & 0.35  & 0.937 & 22.18 \\
SynSin~\cite{Wiles_2020_CVPR}      & 0.52          & 0.824 & 22.45          & 0.55  & 0.853          & 20.86 & 0.65  & 0.819 & 19.92 \\
\hline
HVS-Net (w/o Enhancer)      & 0.18          & 0.986 & 28.54          & 0.19  & 0.926          & 26.24 & 0.20  & 0.910 & 26.25 \\
$\text{HVS-Net}^{\dagger}$     & \textbf{0.14} & \textbf{0.986} & \textbf{28.56} & \textbf{0.17}  & \textbf{0.958} & 27.41 & \textbf{0.20}  & \textbf{0.918} & \textbf{26.47} \\
HVS-Net &
  0.15 &
  \textbf{0.986} &
  28.54 &
  \textbf{0.17} &
  0.955 &
  \textbf{27.45} &
  \textbf{0.20} &
  \textbf{0.918} &
  \textbf{26.47} \\ \bottomrule
\end{tabular}
}
\caption{\textit{Quantitative results on synthetic and real images.} For all datasets, the metrics are averaged across all views. Methods with a $\dagger$ symbol are using dense input depth. Both HVS-Net and $\text{HVS-Net}^{\dagger}$ achieve the best results compared to other view synthesis methods. We observe a slight drop of performance without using the proposed Enhancer module.}
\label{quant_table}
\end{table}

\noindent \textbf{Metrics}. We report the PSNR, SSIM, and perceptual similarity (LPIPS)~\cite{LPIPS} of view synthesis between HVS-Net and other state-of-the-art methods.

\begin{figure}[t]
\centering
  \includegraphics[width=0.99\linewidth]{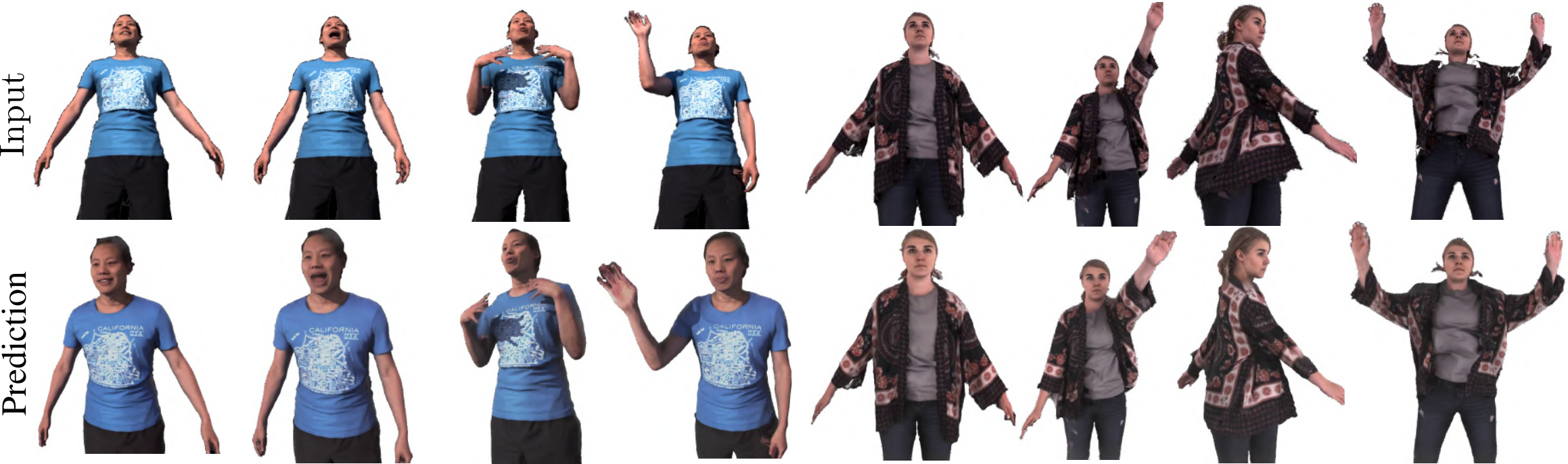}
  \caption{\textit{Generalization to real-world examples.} Our method generalizes well to real-world 4D data and shows robustness w.r.t to different target poses. These results are produced using HVS-Net, trained solely on synthetic data without further fine-tuning.}
\label{fig_qual_robustViews}
\end{figure}
\subsection{Results}\label{Quant_Results}
In Tab.~\ref{quant_table} and Fig.~\ref{fig_qual}, we summarize the quantitative and qualitative results for samples from the RenderPeople dataset.
We first compare the full model HVS-Net against a variant $\text{HVS-Net}^{\dagger}$, which utilizes a dense map as an input.
We observe no significant differences between the predicted novel views produced by HVS-Net when trained using either sparse or dense depth input.
This confirms the effectiveness of the sphere radius predictor: it makes HVS-Net more robust w.r.t. input point cloud density.

In a next step, we evaluate HVS-Net against the current top performing single view human synthesis methods~\cite{LookingGood,Wiles_2020_CVPR}, which do not require per-subject finetuning.
Even though we use dense depth maps as input to $\text{LookingGood}^{\dagger}$~\cite{LookingGood}, the method still struggles to produce realistic results if the target pose deviates significantly from the input viewpoint. 
In the $1^{st}$ row of Fig.~\ref{fig_qual}, $\text{LookingGood}^{\dagger}$~\cite{LookingGood} also struggles to recover clean and accurate textures of the occluded regions behind the hands of the person.
Although both SynSin~\cite{Wiles_2020_CVPR} and HVS-Net utilize the same sparse depth input, the rendered target images are notably different.
SynSin~\cite{Wiles_2020_CVPR} not only performs poorly on the occluded regions but also produces artifacts around the neck of the person, visible in the $2^{nd}$ row of Fig.~\ref{fig_qual}.
In contrast, our method is not only able to render plausible and realistic novel views, but creates them also faithful w.r.t. the input views.
Notice that HVS-Net is able to predict fairly accurate hair for both subjects given very little information.

In a last experiment, we test the generalization ability of our method on real-world 4D data, shown in Fig.~\ref{fig_qual_robustViews}.
Being trained only on synthetic data, this requires generalization to novel identity, novel poses, and bridging the domain gap.
In the 4D scans, the subjects are able to move freely within the capture volume.
We use a fixed, virtual 3D sensor position to create the sparse RGB-D input stream for HVS-Net.
The input camera is placed near the feet of the subjects and is facing up.
As can be seen in Fig.~\ref{fig_teaser} and Fig.~\ref{fig_qual_robustViews}, HVS-Net is still able to perform novel view synthesis with high quality.
Despite using sparse input depth, our method is able to render realistic textures on the clothes of both subjects.
In addition, facial expressions such as opening the mouth or smiling are also well-reconstructed, despite the fact that the static or animated scans used to train our network did not have a variety of facial expressions.
The quality of the results obtained in Fig.~\ref{fig_qual_robustViews} demonstrates that our approach can render high-fidelity novel views of real humans in motion.
We observe that the generated novel views are also temporally consistent across different target view trajectories.
For additional results and video examples, we refer to the supplementary material.

\subsection{Ablation Studies and Discussion}\label{Abs}
\begin{figure}[t]
\centering
  \includegraphics[width=0.99\linewidth]{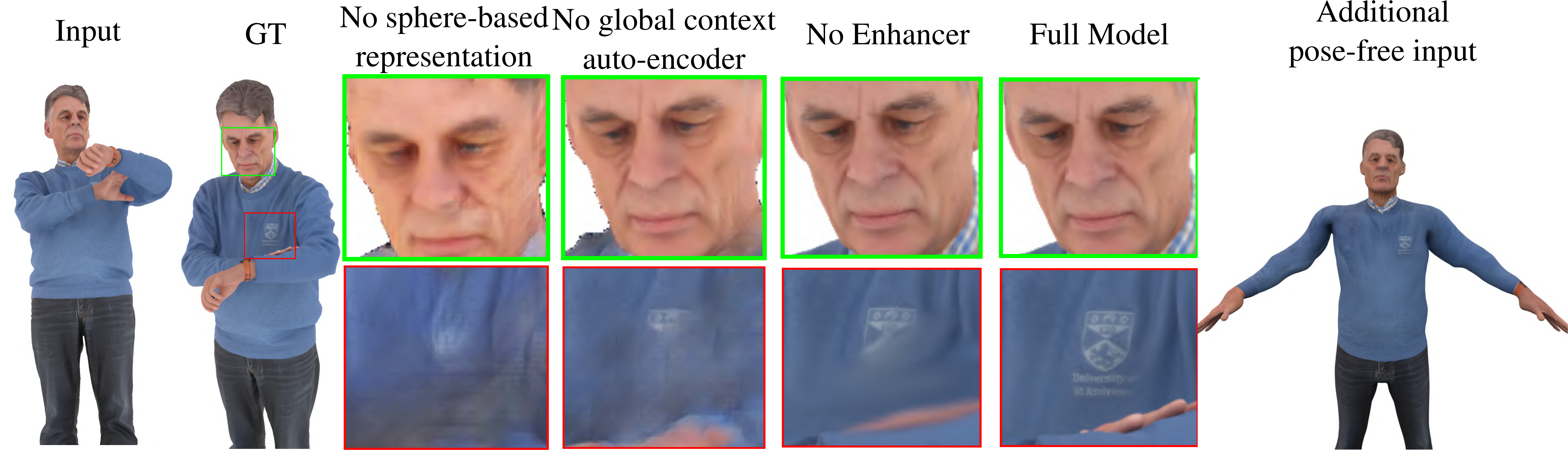}
  \caption{\textit{Qualitative ablation study.} Comparison of the ground-truth with predicted novel views by several variants of the proposed HVS-Net.}
\label{fig_qual_ablation}
\end{figure}

\noindent\textbf{Model Design.}
Tab.~\ref{tab:abl} (left) and Fig.~\ref{fig_qual_ablation} summarize the quantitative and qualitative performance for different model variants on the test set of the RenderPeople dataset~\cite{renderppl}.
HVS-Net without the sphere-based representation does not produce plausible target views (see, for example, the rendered face, which is blurry compared to the full model).
This is due to the high level of sparsity of the input depth, which leads to a harder inpainting problem for the neural network that addresses this task.
Replacing the Fast Fourier Convolution residual blocks of the global context inpainting model with regular convolution layers leads to a drop in render quality in the occluded region (red box).
Using the proposed model architecture, but without the enhancer ($5^{th}$ column of Fig.~\ref{fig_qual_ablation}) leads to a loss of detail in texture. 
In contrast, the full proposed model using the Enhancer network renders the logo accurately.
Note that this logo is completely occluded by the human's hands so it is non-trivial to render the logo using a single input image.

\begin{figure}[t]
\centering
  \includegraphics[width=0.99\linewidth]{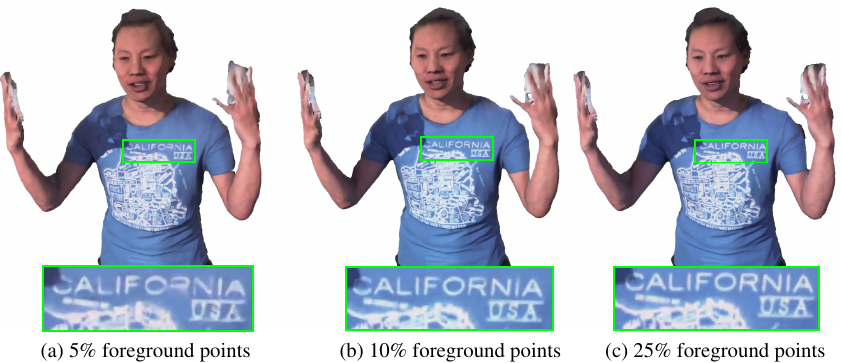}
  \caption{\textit{HVS-Net sparsity robustness.} We randomly sample (a) 5\%, (b) 10\% and (c) 25\% of dense depth points as input depth map and use it as an input for HVS-Net to predict novel views. The text in the T-shirt is reconstructed at high-fidelity with 25\% of the depth points utilized.}
\label{fig_abs_differentDepths}
\end{figure}

\noindent\textbf{Sparse Depth Robustness}.
In Fig.~\ref{fig_abs_differentDepths}, we show novel view synthesis results using different levels of sparsity of the input depth maps.
We first randomly sample several versions of the sparse input depth and HVS-Net to process them.
Our method is able to maintain the quality of view synthesis despite strong reductions in point cloud density.
This highlights the importance of the proposed sphere-based rendering component and the enhancer module.
As can be seen in Tab.~\ref{tab:abl} (right), we observe a slight drop of performance when using 5\% or 10\% of the input maps.
To balance between visual quality and rendering speed, we suggest that using 25\% of the input depth data is sufficient to achieve similar results compared to using the full data.

\setlength{\tabcolsep}{.04cm}
\begin{table}[t]
	\centering
	\footnotesize
	\begin{tabular}[b]{@{}lccc@{}}
        \toprule
        \begin{tabular}[c]{@{}c@{}}Method \\  Variant\end{tabular} & LPIPS$\downarrow$         & SSIM$\uparrow$           & PSNR$\uparrow$           \\ \midrule
        No Sphere Repres.    & 0.22          & 0.934          & 26.15          \\%representation
        No Global Context  & 0.21          & 0.954          & 26.82          \\%reasoning
        No Enhancer                   & 0.18          & 0.967          & 27.92          \\
        HVS-Net (full)              & \textbf{0.15} & \textbf{0.986} & \textbf{28.54} \\ \bottomrule
    \end{tabular}
    \;\;\;\;
	\begin{tabular}[b]{@{}ccccc@{}}
        \toprule
        \begin{tabular}[c]{@{}c@{}}Input depth \\  (\%)\end{tabular} & \begin{tabular}[c]{@{}c@{}}Run-time$\uparrow$\\ (fps)\end{tabular} & LPIPS$\downarrow$ & SSIM$\uparrow$ & PSNR$\uparrow$ \\ \midrule
        5   & \textbf{25} & 0.17          & 0.985          & 28.27          \\
        10  & 22 & 0.15          & 0.986          & 28.54          \\
        25  & 21 & \textbf{0.14} & 0.986 & 28.55          \\
        100 & 20 & 0.14          & 0.986          & \textbf{28.56} \\ \bottomrule
    \end{tabular}
	\caption{\textit{\textbf{Left}:  Ablation study.} Reconstruction accuracy on the RenderPeople testing set. \textit{\textbf{Right}: Reconstruction accuracy and inference speed} using different levels of input depth sparsity.}
	\label{tab:abl}
\end{table}

\noindent \textbf{Inference Speed}.
For AR/VR applications, a prime target for a method like the one proposed, runtime performance is critical.
At test time, HVS-Net generates \(1024\times1024\) images at 21FPS using a single NVIDIA V100 GPU.
This speed can be further increased with more efficient data loaders and an optimized implementation that uses the NVIDIA TensorRT engine.
Finally, different depth sparsity levels do not significantly affect the average runtime of HVS-Net, which is a plus compared to prior work. 

\section{Conclusion}\label{Conclusions}

We presented HVS-Net, a method that performs novel view synthesis of humans in motion given a single, sparse RGB-D source.
HVS-Net uses a sphere-based view synthesis model that produces dense features of the target view; these are then utilized along with an autoencoder to complete the missing details of the target viewpoints.
To account for heavily occluded regions, we propose an enhancer module that uses an additional unoccluded view of the human to provide additional information and produce high-quality results based on an novel IUV mapping.
Our approach generates high-fidelity renders at new views of unseen humans in various new poses and can faithfully capture and render facial expressions that were not present in training. 
This is especially remarkable, since we train HVS-Net only on synthetic data; yet it achieves high-quality results across synthetic and real-world examples. 

\noindent\textbf{Acknowledgements}:
The authors would like to thank Albert Para Pozzo, Sam Johnson and Ronald Mallet for the initial discussions related to the project. % their Metamates 

\clearpage
\bibliographystyle{splncs04}
\bibliography{egbib}
\end{document}

\pagestyle{headings}
\mainmatter

\title{Free-Viewpoint RGB-D Human Performance \\Capture and Rendering} 

\renewcommand\thesection{\Alph{section}}
\renewcommand\thesubsection{\thesection.\arabic{subsection}}

\definecolor{down}{RGB}{255, 128, 128}
\definecolor{up}{RGB}{0, 128, 128}
\definecolor{orange}{RGB}{255, 128, 0}
\titlerunning{Free-Viewpoint RGB-D Human Performance Capture and Rendering} 
\author{Phong Nguyen-Ha \inst{1*} \and
Nikolaos Sarafianos \inst{2}\and \\
Christoph Lassner \inst{2} \and Janne Heikkil\"a \inst{1} \and
Tony Tung \inst{2}}
\authorrunning{P. Nguyen et al.}
\institute{Center for Machine Vision and Signal Analysis, University of Oulu, Finland
\and
Meta Reality Labs Research, Sausalito\\ \href{https://www.phongnhhn.info/HVS_Net}{https://www.phongnhhn.info/HVS\_Net}} 
\maketitle

\begin{figure}[ht]
\centering
  \includegraphics[width=0.98\linewidth]{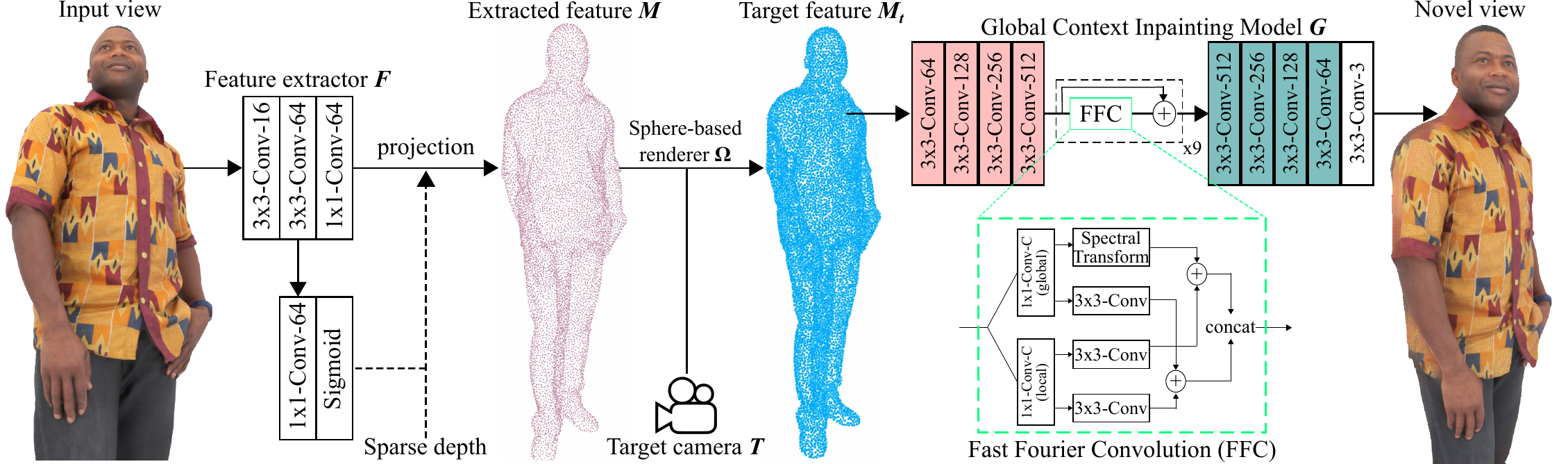}
  \caption{\textit{Detailed architecture of the sphere-based view synthesis network.} The feature extractor $F$ first use three convolution layers with stride 1 to extract the features of the input view. We then infer the radius of each sphere by passing the learned features through another convolution layer and the sigmoid activation function. The \textcolor{up}{green} and \textcolor{down}{red} convolution layers of $G$ module scale up and down the feature maps respectively.}
\label{fig_S}
\end{figure}

In this supplementary material we provide additional details regarding our network designs (Sec.~\ref{Network_design}), as well as implementation details (Sec.~\ref{Implementation}). Additional qualitative evaluations and results are shown in the supplemental video. Finally, we discuss the limitation of our approach (Sec.~\ref{Limit}). 
\blfootnote{*This work was conducted during an internship at Meta Reality Labs Research.}

\section{Network Designs}\label{Network_design}

In this section, we describe the technical details of two sub-networks of our proposed HVS-Net: a sphere-based view synthesis $S$ and a enhancer model $E$.

\subsection{Sphere-based view synthesis model $S$}
\begin{figure}[t]
\centering
  \includegraphics[width=0.95\linewidth]{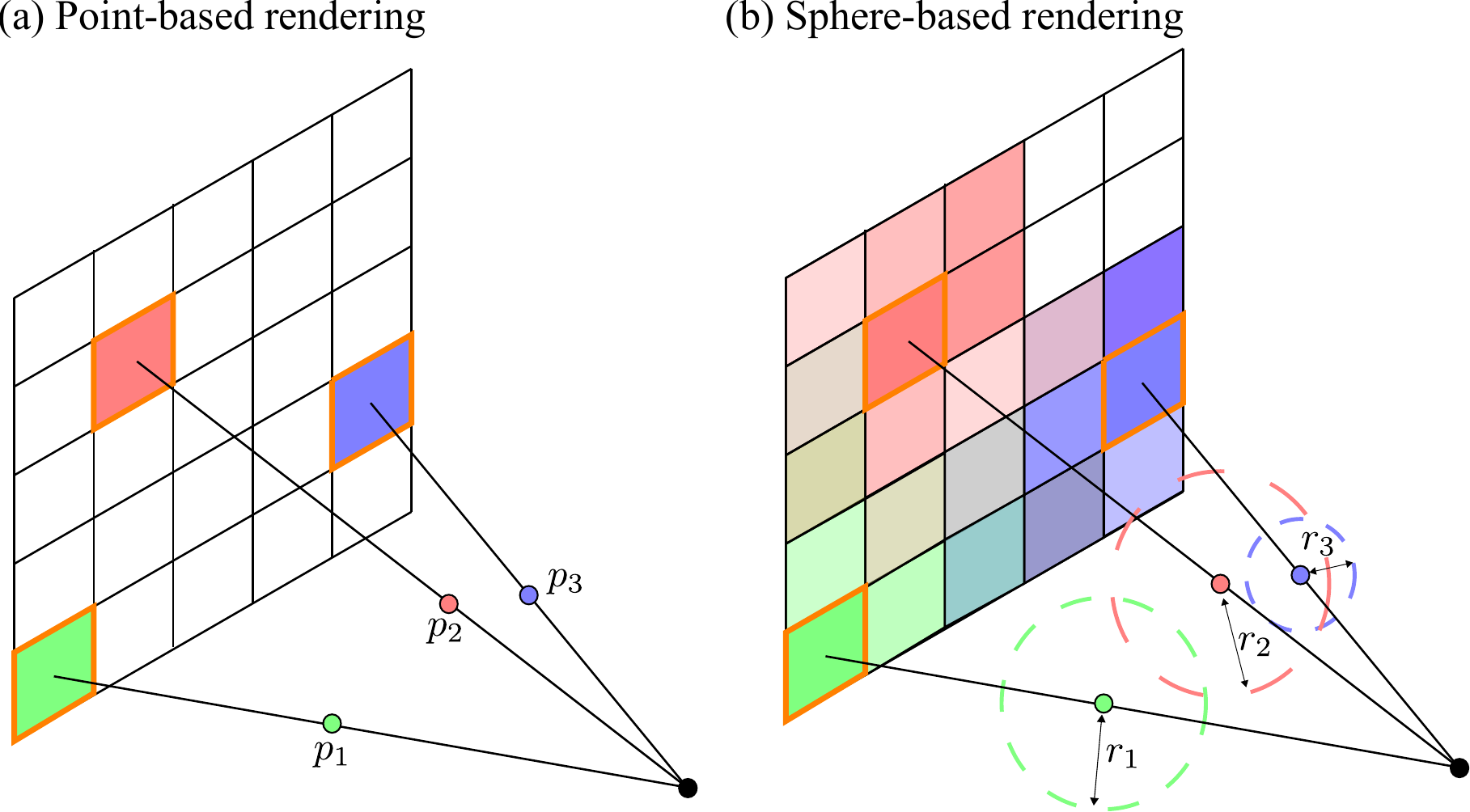}
  \caption{Visualization of the \textit{rendered features} between (a) point and (b) sphere-based rendering methods. Point-based method~\cite{Wiles_2020_CVPR} can only render pixels (\textcolor{orange}{orange} boxes) that have valid 3D coordinates. In contrast, sphere-based method~\cite{lassner2021pulsar} uses learned radius $r_i$ of each point $p_i$ to render neighboring pixels which leads to a denser feature map.}
\label{fig_pulsar}
\end{figure}

\noindent\textbf{Sphere-based feature warping}.
The architecture of the sphere-based view synthesis model $S$ is shown in Fig.~\ref{fig_S}. 
Instead of directly rendering novel views using the RGB input image, we first passed it through a feature extractor $F$ which consists of three convolution layers with stride 1 to maintain the spatial resolution. We choose the features $f_i$ as the values of $M$ where there is a valid depth value. We estimate per-sphere radius $r_i$ by passing $M$ to another convolution layer with sigmoid activation function. In Fig.~\ref{fig_pulsar}, we show the visualization of rendered feature maps from a set of sparse points using point and sphere-based renderers. In case of point-based rendering~\cite{Wiles_2020_CVPR}, each 3D point ${p_i}$ can render a single pixel. Therefore, a large amount of pixels can not be rendered because there is no ray connecting those pixels with valid 3D points.
In contrast, the sphere-based neural renderer~\cite{lassner2021pulsar} $\Omega$ renders a pixel by blending the colors of any intersected spheres with the given ray. Since we estimate radius $r_i$ of each sphere (dashed circle) using a shallow network, this allows us to render pixels that do not have a valid 3D coordinates. 
As a result, we obtain a much denser feature maps as can be seen in the Fig. 2 of the main paper. Note that, $\Omega$ is fully-differentiable and renders target feature maps very efficiently using PyTorch3D~\cite{PT3D}.

\noindent\textbf{Global context inpainting model}. We render the novel view using a global context inpainting model $G$. We design the architecture of the $G$ module based on the encoder-decoder structure with skip connections and nine residual blocks are also utilized in the bottleneck.

In each residual block, we replace the regular convolution layers with the recently proposed Fast Fourier Convolution(FFC)~\cite{FFC} which possesses the non-local receptive fields. According to the spectral convolution theorem in Fourier theory, point-wise update in the spectral domain globally affects all features involved in the Fourier transform. The FFC layer splits the input features into local and global branches. The local branch utilizes conventional convolution layers to obtain local features. In contrast, the global branch includes a Spectral Transform block~\cite{LAMA} which uses channel-wise Fast Fourier Transform~\cite{FFT} to enable image-wide receptive field. The output of both branches are then summed, aggregated before adding to the residuals.

\noindent\textbf{Outputs}. The view synthesis model $S$ not only predicts an RGB image $I_p$ of the target view but also a foreground mask $I_m$ and a confidence map $I_c$. We employ three different $3 \times 3$ convolution layers to predict those outputs using the output of the final layer of the $G$ module. Thus, we apply the predicted foreground mask and confidence map to the predicted novel image as follow: $I_p = I_p \times I_m \times I_c$.
We train the model $S$ using the photometric loss 
$\mathcal{L}_{photo}$ as defined in the main paper.

\begin{figure}[t]
\centering
  \includegraphics[width=0.8\linewidth]{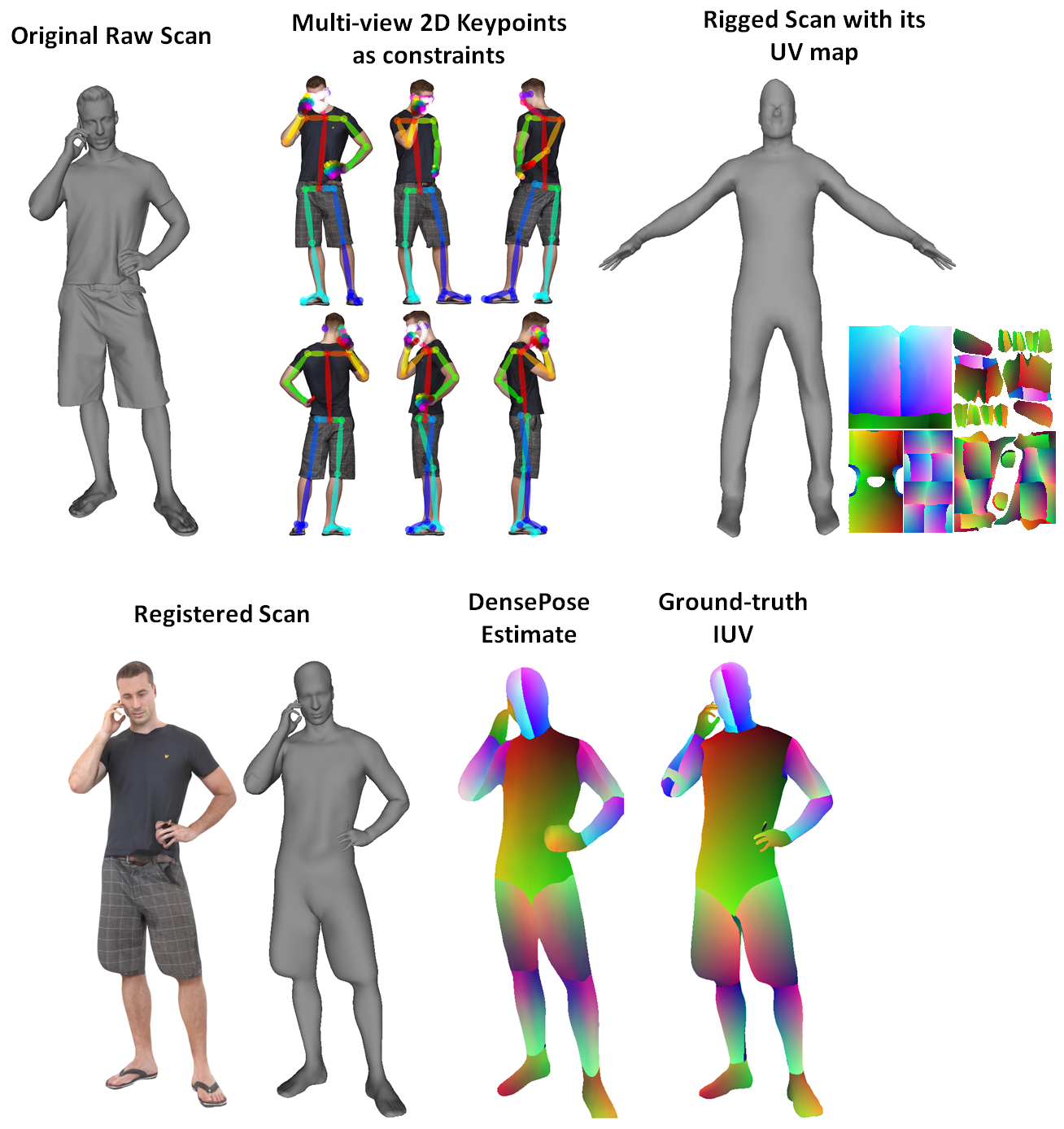}
  \caption{\textit{Process for IUV ground-truth generation} Given a raw synthetic scan of a clothed human (top left) we perform non-rigid registration with 2D keypoints as additional constraints (top-middle) and obtain the registered scan to the body template (bottom left) and the rigged scan (top right) which is animation ready. Using the corresponding UV map we can now obtain accurate IUV ground-truth (bottom right) that we use to train the proposed HD-IUV model. We provide the corresponding DensePose estimate to demonstrate the stark difference between the two in terms of quality as well as coverage.}
\label{fig_iuv_gt}
\end{figure}

\subsection{Enhancer model $E$}
\noindent\textbf{Ground-truth Data}:
We use the RenderPeople dataset~\cite{renderppl} to train all our models; which comprises of 1000 watertight raw meshes. To obtain IUV ground-truth we first fit an SMPL-like parametric body model to the scans and then perform non-rigid registration for all meshes and rig them for animation. In that way we obtain 1000 rigged models to which we can apply the same IUV map during rendering with an emission shader in Blender Cycles and thus obtain per-pixel perfect IUV ground-truth given an RGB input. This process is depicted in Fig.~\ref{fig_iuv_gt}.

\begin{figure}[t]
\centering
  \includegraphics[width=0.99\linewidth]{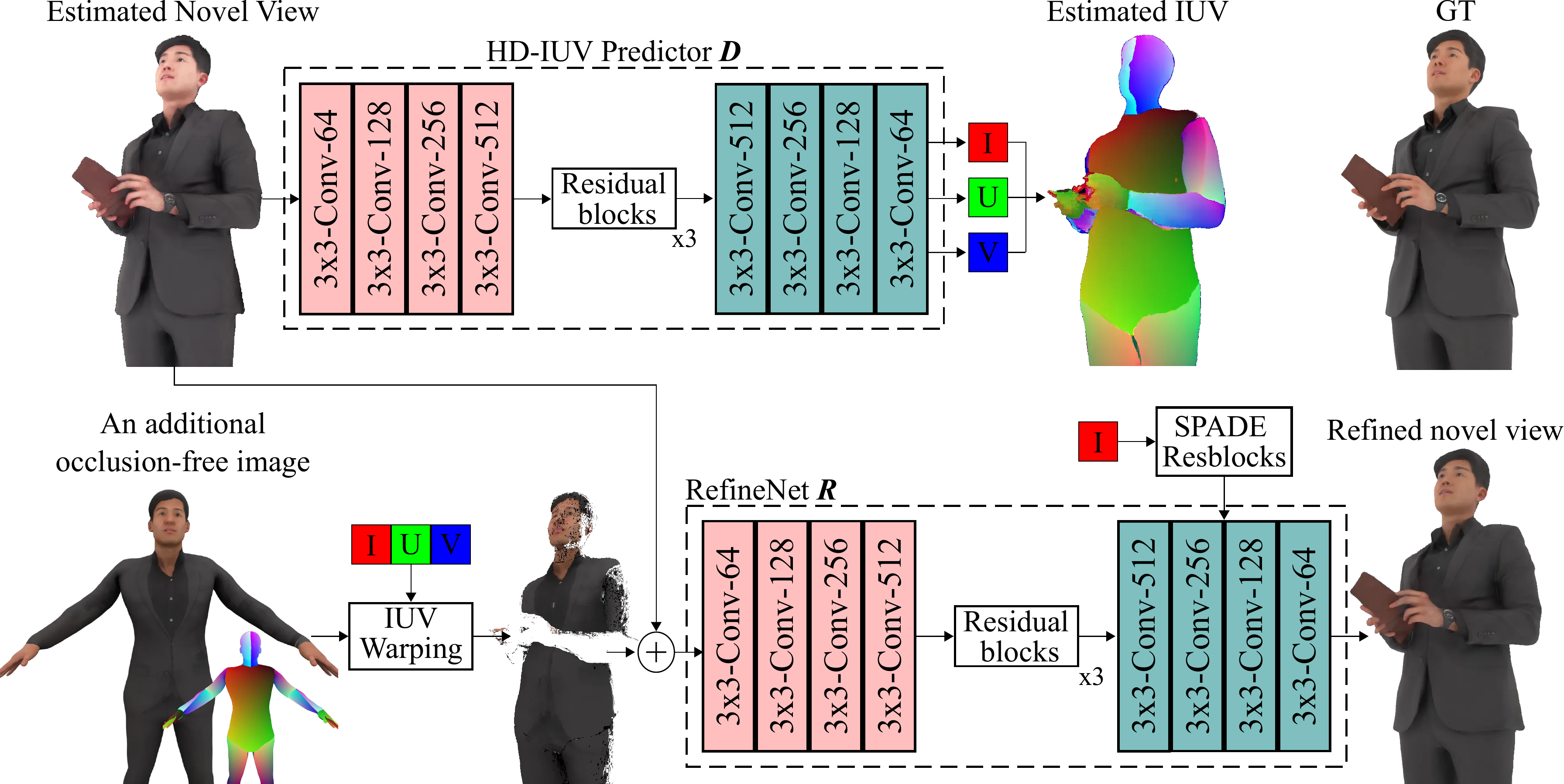}
  \caption{\textit{IUV-based image refinement.} Using an additional occlusion-free input, we refine the initial estimated novel view by training the Enhancer $E$ network. We infer the dense correspondences of both predicted novel view and occlusion-free image using a novel \textit{HD-IUV} module. The occlusion-free image is warped to the target view and then refined by an auto-encoder. The refined novel view shows better result on the occluded area compared to the initial estimated.}
\label{fig_DR}
\end{figure}

\noindent\textbf{HD-IUV predictor $D$}: Now that we have generated pairs of RGB images and ground-truth IUV maps the next step is to train a network that given an RGB image of a human, can establish accurate per-pixel correspondences \textbf{for each pixel} corresponding to the clothed human (see Fig.\ref{fig_DR}). Note that the key difference between this approach and what methods such as DensePose~\cite{alp2018densepose} or CSE~\cite{neverova2020continuous} are doing which is dense correspondence estimates to the unclothed human body. In addition because most approaches are trained on the DensePose-COCO dataset~\cite{alp2018densepose} which comprises sparse (only \mytilde100 discrete points per human) and noisy annotations such predictions are usually inaccurate and not applicable to our application that targets clothed humans. This is also depicted in Fig.~5 of the main paper where its clear that DensePose IUV estimates result into poor texture warpings. 

To train our model which we term as HD-IUV (that stands for High-Definition IUV) we employed an encoder-decoder architecture with four \textcolor{down}{downsampling} and \textcolor{up}{upsampling} convolution layers along with skip connections between them while the bottleneck comprises 3 residual blocks. This design is justified by the fact that our input-output pairs are always well aligned due to the dense correspondences established by HD-IUV which is not the case with prior work. For HD-IUV, we utilize instance  normalization~\cite{ulyanov2016instance} and the ReLU activation function in all layers of the network besides the 3 output branches for each task ($I$, $U$, $V$ outputs). The UV branches have \(256\) output channels (since the UV predictions can take any possible value), whereas the $I$ channel has \(25\) channels which correspond to 24 body parts and background. In all branches a \(1\times\ 1\) convolution is applied and its output is an unnormalized logit that is then fed to the cross-entropy losses. 
Each task's scores are fed to their respective classification losses which are used to train the network as: 
\begin{equation}
L_{IUV} = \lambda_{I} * L_I + \lambda_U * L_U + \lambda_V * L_V 
\end{equation}
where \(\lambda_{i}, L_i\) are the respective weighting parameters and loss functions for the \(I, U, V\) channels. Framing this problem as a multi-task learning problem (3 tasks) where the \(U, V\) and \(I\) tasks are \((256D, 256D, 25D)\) per-pixel classification problems respectively, ended up being a very effective approach to enforce strong supervisions for the surface correspondences that other losses we experimented with could not achieve. In addition we employed a silhouette loss to ensure that dense correspondence estimates are provided for each pixel of the foreground clothed human.
Finally, using the predicted IUV, we can warp the occlusion-free input image to the target camera using the texture transfer technique\footnote{\href{https://github.com/facebookresearch/DensePose/blob/main/notebooks/DensePose-RCNN-Texture-Transfer.ipynb}{Texture Transfer Using Estimated Dense Coordinates}} from DensePose~\cite{alp2018densepose}.

\noindent\textbf{Refinement module}
In this section, we utilize the warped image $I_w$ from previous step 
to enhance the initially estimated target view $I_p$ using a refinement module $R$. 
Based on the predicted confidence of the view synthesis network, we combine both images as follows: $\hat{I} = I_p + (1 - I_c) * I_w$ where $\hat{I}$ is fed to a encoder-decoder network for the refinement purposes.
In this work, we try to generate humans at the novel viewpoints so rendering realistic human body parts is required. We observe that the predicted semantic $I$ contains valuable information about the semantic information of the human in the target camera. Therefore, we use the SPADE normalization~\cite{park2019SPADE} to inject the semantics $I$ to the decoder of the refinement module. As can be seen in the qualitative results, the refined image is photo-realistic compared to the ground-truth image. Note that, we use the same discriminator with ~\cite{park2019SPADE} to perform adversarial training between both before and after refined images and the ground-truth novel views.

\noindent\textbf{Discussion} Here we discuss the effectiveness of our proposed HD-IUV over DensePose~\cite{alp2018densepose} representations to refine the target views. As can be seen in the Fig.~8 of the main paper, our Enhancer model can handle heavy occlusions using just a single photo. We emphasize that the HD-IUV representation is crucial for this refinement step because we can obtain pixel-aligned warped images at the target viewpoints compared to the ground-truth data. Therefore our warped images have higher quality compared to those produced by DensePose. 

\section{Implementation Details}\label{Implementation}
The models were trained with the Adam optimizer using a 0.004 learning rate for the discriminator, 0.001 for both the view synthesis model $R$ and the enhancer module $E$ and momentum parameters (0, 0.9). The input/output of our method are $1024 \times 1024$. We implement HVS-Net in PyTorch and the training across our large-scale dataset with all identities and views took 2 days to converge on 4 NVIDIA V100 GPUs.

\section{Limitations}\label{Limit}
Despite producing appealing results on real-world data, the proposed method is trained solely on synthetic data.
It manages to bridge the domain gap remarkably well, however we believe its performance could be further improved by integrating real-world data into the training set.

However, gathering such data is not trivial: generating (close to) noise-free point clouds for training requires elaborate multi-view capture systems, possibly enhanced with controlled lighting to simulate varying lighting conditions.
A way to circumvent this partially is to train on a large-scale synthetic dataset~\cite{wood2021fake} and then fine-tuning on a smaller-scale real-world dataset.
This, at least, reduces the amount of data that has to be captured.

Another limitation we identified is that the warped image used as input to the enhancer model has lower quality compared to the initial estimated novel view.
This is independent of the quality of the IUV mapping and is an inherent problem of the differentiable warping operation.
Improving this operation could be a promising direction for future work that could increase the upper bound in quality for the novel view synthesis of fine structures in occlusion scenarios.

\bibliographystyle{splncs04}
\bibliography{egbib}